\documentclass[conference]{IEEEtran}
\IEEEoverridecommandlockouts
\usepackage{blindtext}
\usepackage{geometry}
\usepackage{cite}
\usepackage{amsmath,amssymb,amsfonts}
\usepackage{algorithmic}
\usepackage{graphicx}
\usepackage{textcomp}
\usepackage{xcolor}
\usepackage{algorithm2e}
\def\BibTeX{{\rm B\kern-.05em{\sc i\kern-.025em b}\kern-.08em
    T\kern-.1667em\lower.7ex\hbox{E}\kern-.125emX}}
\begin{document}
\newcommand{\INDSTATE}[1][1]{ \STATE\hspace{#1\algorithmicindent}}

\title{Improving the Real-Data Driven Network Evaluation Model for Digital Twin Networks}

\author{
\IEEEauthorblockN{Hyeju Shin, Ibrahim Aliyu, Abubakar Isah, Jinsul Kim*}
\IEEEauthorblockA{Dept. of ICT Convergence System Engineering, Chonnam National University, South Korea \\
\{sinhye102, aliyu, abubakarisah, jsworld*\}@jnu.ac.kr}
}

\maketitle

\begin{abstract}
With the emergence and proliferation of new forms of large-scale services such as smart homes, virtual reality/augmented reality, the increasingly complex networks are raising concerns about significant operational costs. As a result, the need for network management automation is emphasized, and Digital Twin Networks (DTN) technology is expected to become the foundation technology for autonomous networks. DTN has the advantage of being able to operate and system networks based on real-time collected data in a closed-loop system, and currently it is mainly designed for optimization scenarios. To improve network performance in optimization scenarios, it is necessary to select appropriate configurations and perform accurate performance evaluation based on real data. However, most network evaluation models currently use simulation data. Meanwhile, according to DTN standards documents, artificial intelligence (AI) models can ensure scalability, real-time performance, and accuracy in large-scale networks. Various AI research and standardization work is ongoing to optimize the use of DTN. When designing AI models, it is crucial to consider the characteristics of the data. Since communication networks are composed of a graph structure connecting distributed devices, reflecting this spatial information when training AI models will help improve accuracy. This paper presents an autoencoder-based skip connected message passing neural network (AE-SMPN) as a network evaluation model using real network data. The model is created by utilizing graph neural network (GNN) with recurrent neural network (RNN) models to capture the spatiotemporal features of network data. Additionally, an AutoEncoder (AE) is employed to extract initial features. The neural network was trained using the real DTN dataset provided by the Barcelona Neural Networking Center (BNN-UPC), and the paper presents the analysis of the model structure along with experimental results.
\end{abstract}


\section{Introduction}
With the advancement of the Internet of Things (IoT), cloud computing, and 5G technology, new forms of large-scale services such as smart homes, autonomous vehicles, and virtual reality/augmented reality (VR/AR) are emerging and proliferating\cite{a}.
As billions of users of these large-scale services become connected to numerous devices and applications, there is concern that the complex network will require significant operational costs\cite{b}. Managing complex networks requires more human resources with specialized knowledge, resulting in higher operating costs. To address this issue, the need for network management automation is emphasized\cite{c}\cite{d}.

\begin{figure}[htbp]
\centerline{\includegraphics[width=9cm]{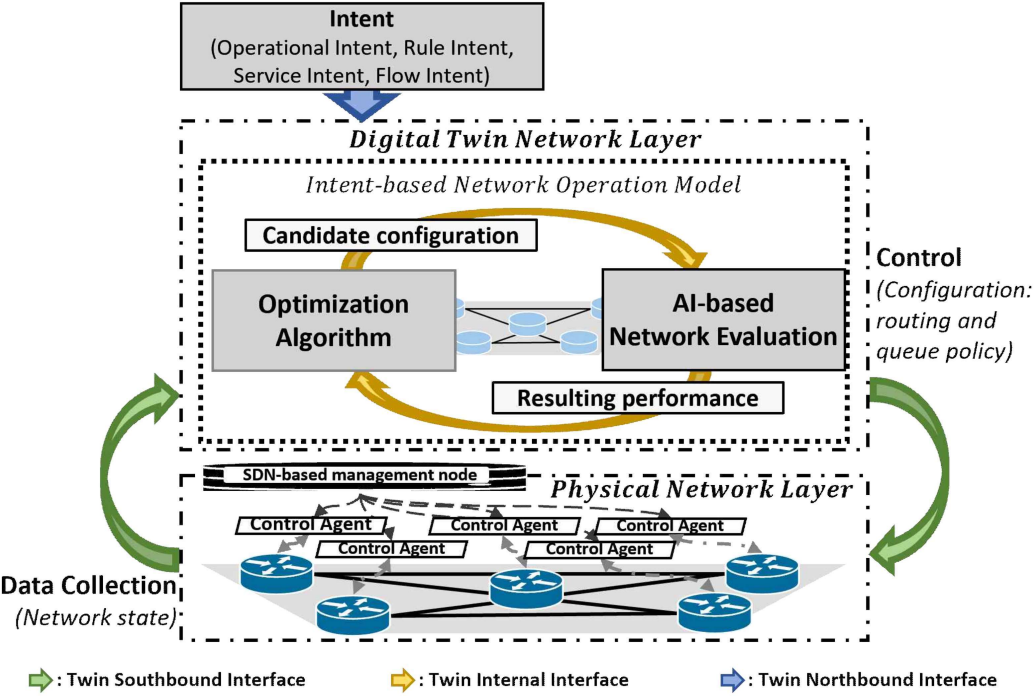}}
\caption{Example of the optimization scenario in DTN.}
\label{fig1}
\end{figure}

Digital Twin (DT) is a technology that represents physical objects in the digital world in real-time. The main advantage of DT is its ability to enable real-time interaction and automated process optimization based on the connection between virtual and physical realities\cite{e}. As a result, it is being adopted in various fields\cite{t-1}\cite{E-I}. Recently, a technology called Digital Twin Network (DTN) has emerged as a DT technology for communication networks, and it is expected to become the foundation technology for autonomous networks\cite{h}. Compared with traditional network management systems, DTN supports bidirectional interaction between virtual and real closed-loop systems, enabling network operation and management based on real-time collected data\cite{o-1}. In particular, as shown in Fig. 1, current DTN technology research is mainly designed for optimization scenarios\cite{irtf-draft}\cite{E-h}. In this case, the development of a network performance evaluation model based on real data is considered important\cite{routenet-f}.

To improve network performance in an automated physical network optimization scenario, it is necessary to select appropriate configuration values. Therefore, it should be possible to evaluate the network performance when the configuration information is updated. Considering that the selected optimal configuration information can be automatically deployed to the physical network, it is important to accurately evaluate the configuration information based on real data rather than simulation data\cite{crnt arx}. However, most of the current research on network evaluation models is conducted using simulation data\cite{routenet}\cite{lst conf}\cite{ath ppr}. Therefore, in this paper, we conducted the training and validation of network evaluation models using real-world DTN datasets collected from the physical testbed provided by \cite{data}.
It is crucial to select a model that suits the data's characteristics when training an artificial intelligence (AI) model. If dealing with communication network data presented in a graph structure, it is advisable to use graph neural network (GNN) models that can represent such spatial traits\cite{routenet-f}\cite{gnn-draft}.

Accordingly, this paper proposed an autoencoder-based skip-connected message-passing neural network (AE-SMPN) model. The model is based on the gnn series routenet\cite{routenet}. The ability of the autoencoder (AE) to reduce dimensions is exploited for accurate initial feature extraction, and skip connections compensate for the problem of vanishing gradients during training. The main contributions of this study are outlined below:
\begin{itemize}
\item Present the DTN optimization scenario along with the architecture and propose a data pipeline related to the model training process.
\item Train and validate a proposed network evaluation model using real DTN datasets.
\item Propose an AE-SMPN model that includes an AE-based embedding process to ensure the accuracy of the initial feature extraction of the network data.
\end{itemize}

\subsection{Related Work}

\begin{figure}[htbp]
\centerline{\includegraphics[width=8.5cm]{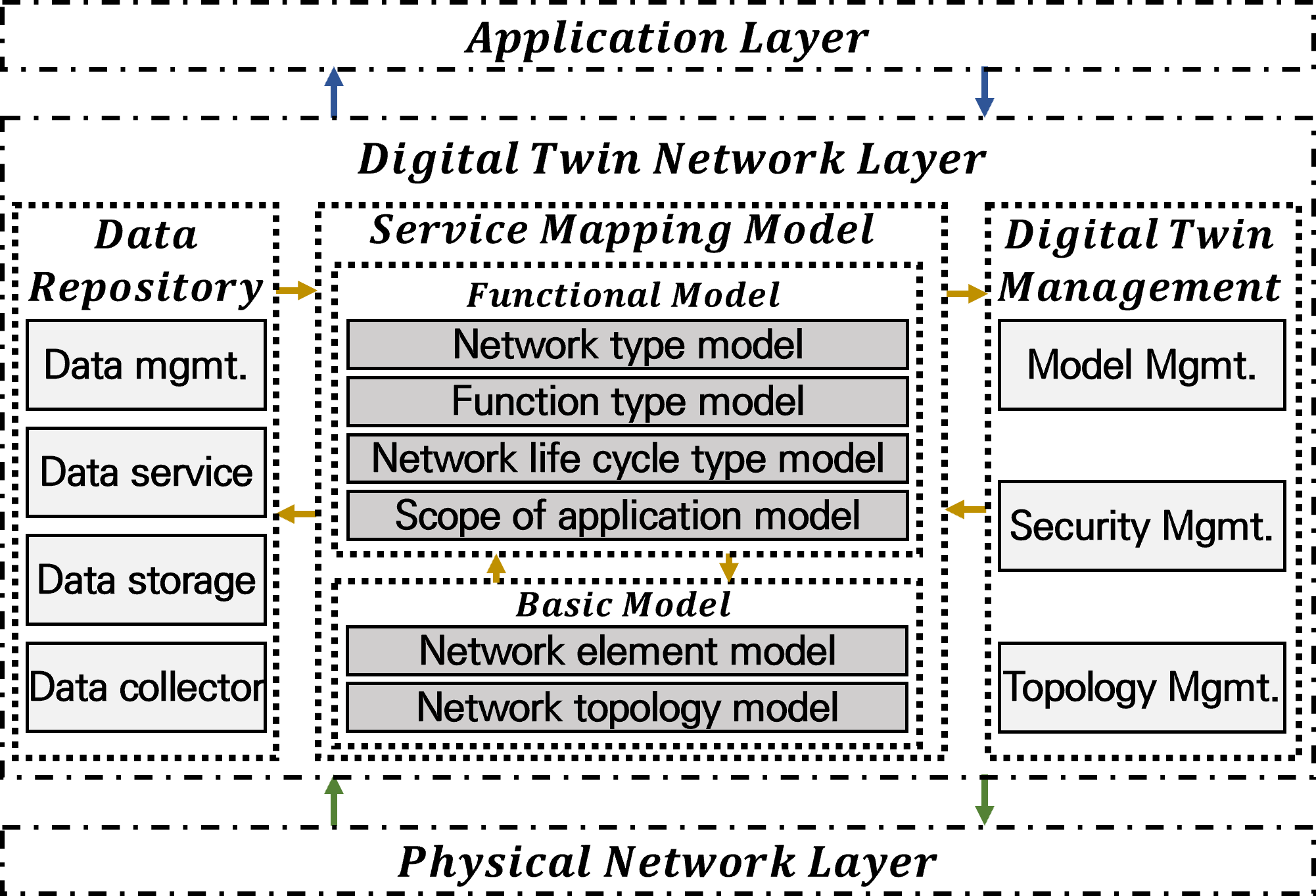}}
\caption{DTN basic architecture.}
\label{fig2}
\end{figure}

DTN is a DT technology for communication network systems and is gaining attention as a technology for network automation. As DTN technology research is in its early stages, the definition of concepts and architecture is important. Therefore, various DTN standardization efforts on different topics are actively underway \cite{irtf-draft}\cite{gnn-draft}. In this section, the basic architecture of DTN and its subsystems are explained along with the functions they provide.

DTN consists of three layers, as shown in Fig. 2, and can achieve network lifecycle management through the interaction of each layer. First, the physical network layer contains the network to be managed by DTN. It is regarded as a data source. Based on the collected data, appropriate user intent solutions can be deployed in the digital twin network layer to control the physical network. Second, the application layer provides the user requirements, including intent, to the digital twin network layer, enabling the operation and management of the physical network according to the requirements.
Lastly, the digital twin network layer offers various models to operate and manage the network, using network state data collected from the physical network layer and user intent information obtained from the application layer. The digital twin network layer comprises three subsystems: a data repository that collects and provides data, DTN management that oversees the digital twin network layer, and a service mapping model that offers solutions suitable for scenarios based on the received data. The section intimately associated with machine learning models for network automation is the service mapping model, and the intent-based SLA/QoS optimizer model in Fig. 1 serves as an example.

\subsection{Organization}
The rest of the paper is organized as follows. In Section II, the system model is presented, and in Section III, the proposed method based on the network evaluation algorithm is explained. In Section IV, the experimental results are provided and analyzed, and finally, in Section V, the conclusion is described along with future research.

\section{System Model}
Fig. 3 shows the learning pipeline of the network evaluation model in the DTN structure. The network data collected from the physical network is stored in a database designed considering the characteristics of the data, and the stored data is embedded in a structure suitable for model learning and provided to the ML model.

\begin{figure}[htbp]
\centerline{\includegraphics[width=8.5cm]{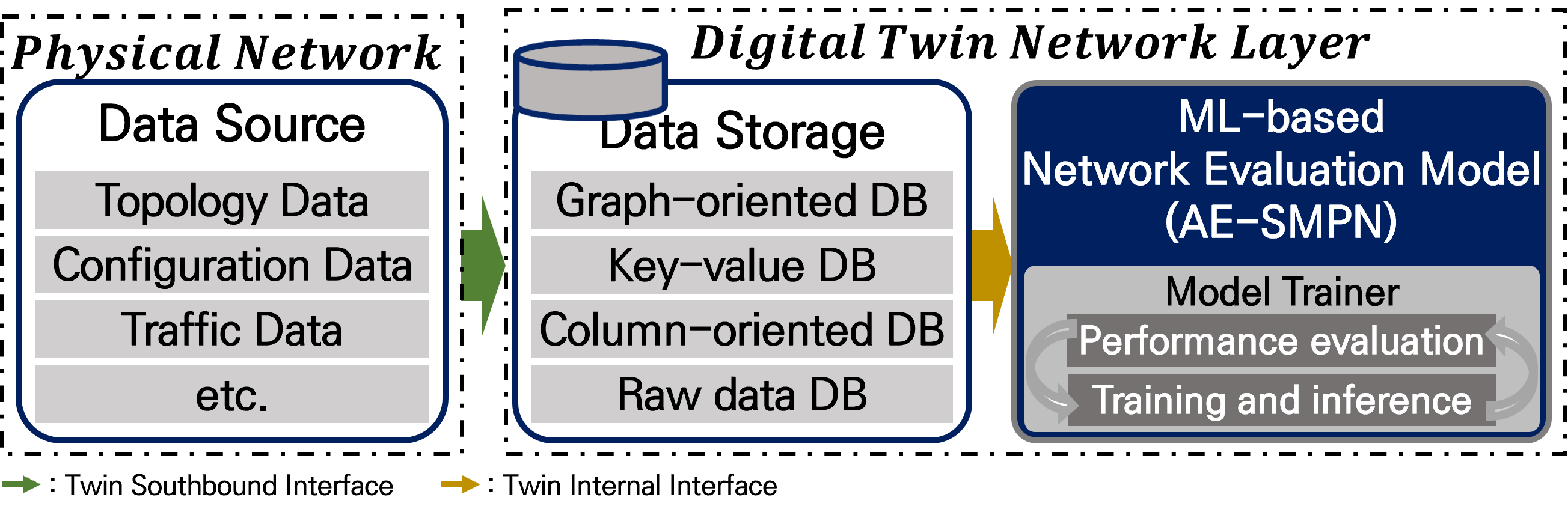}}
\caption{ML model training pipeline in DTN.}
\label{fig3}
\end{figure}

\subsection{AutoEncoder (AE)}
We provide a detailed explanation of the AE model used to extract the feature from network data. AE is a neural network that compresses incoming input data as much as possible and then restores the compressed data to its original input form. As the input data passes through the encoder network, it is compressed and the latent vector value is derived. The equation for a linear encoder is as follows:

\begin{equation}
z=h(x)=W_{e}x+b_e\label{eq1}
\end{equation}

The compressed z-vector passes through the decoder network to produce an output value of the same size as the input data, and the equation for this is:

\begin{equation}
y=g(z)=W_{d}z+b_d\label{eq2}
\end{equation}

For model training, the L2 loss, which is the difference between the input data x and the output data y after passing through the AE, can be used as the loss function. The equation is as follows:

\begin{equation}
L_{AE}=\sum_{x \in D}L_2(x,y)=\sum_{x \in D}{||x-y||^2}\label{eq3}
\end{equation}

where the x and y dimensions are the same size.

\subsection{Graph Neural Network (GNN)}
The GNN model is a neural network suitable for graph-structured data. Graph-structured data consists of nodes and edges, and the connection relationship between nodes is considered an important feature. Since communication networks have a graph structure with distributed devices connected, GNN models have been actively used in the field of communication networks recently\cite{routenet-f}\cite{routenet}\cite{gnn-draft}.
The message-passing neural network (MPNN), which is a type of graph neural network, can reflect the spatial information of the entire data by receiving information from neighboring nodes based on the connection information. Each node derives its next hidden state by considering its own hidden state, the hidden state of neighboring nodes, and the connection information. This process is called the message-passing phase, and the formula for deriving the message of node v is as follows:

\begin{equation}
m_v^{k+1}=\sum_{w \in N(v)}M_k(h_v^k,h_w^k,e_{vw})\label{eq4}
\end{equation}

where, $h_{v}$ is the hidden state for node $v$, $h_{w}$ is the hidden state of the neighboring node $w$, and $e_{vw}$ is the link information between nodes.

Based on the derived message, we can calculate the next hidden state for node $v$. This process is called the update phase. In this phase, the current hidden state of node $v$, $h_{v}^{k}$, and the computed message, $m_{v}^{k+1}$, are taken into account. The equation is as follows:

\begin{equation}
h_v^{k+1}=U_k(h_v^k,m_v^{k+1})\label{eq5}
\end{equation}

where, $h_{v}^{0} = x_{v}$, and $U_{k}$ and $M_{k}$ are functions with learning parameters.

The hidden state $h_{v}^{k}$ derived in this way is used to predict the label, and this is called the readout stage. The equation is as follows:

\begin{equation}
\hat{y}=R(\{h_{v}^{K}|v \in G\})\label{eq6}
\end{equation}

where, $R$ is a function with learning parameters.
By repeating the above three processes $K$ times, the $K$th hidden state, $h_{v}^{k}$, for each node $v$ $(v \in G)$ can be derived, and finally the desired value can be predicted by the readout stage.

\subsection{Long Short-Term Memory (LSTM)}
The existing recurrent neural network (RNN) model may encounter the gradient vanishing problem when the length of the input data becomes longer. LSTM is a model designed to solve this problem, and it has a memory cell that can remember previous information for a long time, which allows it to process long sequence data. Before selecting the information to be stored in the memory cell, it determines which information to forget. This process is performed by the Forget Gate, and the equation is as follows:

\begin{equation}
f_t=\sigma(W_f\cdot[h_{t-1},x_t]+b_f)\label{eq7}
\end{equation}

where, $\sigma$ is a sigmoid function that produces numbers between 0 and 1.

To store the necessary information from the input data in the memory cell, an input gate is used. At this time, a tanh layer is used to determine a candidate vector for new information, and the sigmoid function of the input layer can decide which information to store from the candidates. The equation is as follows:

\begin{equation}
i_t=\sigma(W_i\cdot[h_{t-1},x_t]+b_i)\label{eq8}
\end{equation}
\begin{equation}
\tilde{C_t}=tanh(W_C\cdot[h_{t-1},x_t]+b_C)\label{eq9}
\end{equation}

Using the three formulas above, storing the necessary information in the memory cell is possible. The formulas are as follows:

\begin{equation}
C_t=f_t*C_{t-1}+i_t*\tilde{C_t}\label{eq10}
\end{equation}

Finally, the output gate applies a sigmoid function to the input value to determine which value will be generated as the output value from the calculated cell state. This resulting value is then passed through a hyperbolic tangent (tanh) function to retain only the necessary information as the output value for the next cell. The equation is as follows:

\begin{equation}
o_t=\sigma(W_o\cdot[h_{t-1},x_t]+b_o)\label{eq11}
\end{equation}
\begin{equation}
h_t=o_t*tanh(C_t)\label{eq12}
\end{equation}

\section{Problem Formulation}
In this section, we introduce the autoencoder-based skip-connected message-passing neural network (AE-SMPN), which is a network evaluation model based on real network data. This was done as part of the ITU Challenge ``ITU-ML5G-PS-007: Creating a Network Digital Twin with Real Network Data"\cite{itu challenge} and focused on addressing two problems with the existing state-of-the-art model, RouteNet.

\subsection{AE-based Feature Extraction}
The initial feature is one of the important factors in the performance of the ML model. The proposed method uses an autoencoder to extract appropriate initial features for model training based on the main characteristics of the captured packets. If $F$ represents the flow data of the packets, $L_2$ represents the data for layer 2 of the network, and $L_3$ represents the data for layer 3 of the network, each data is embedded into the autoencoder as $AE_f$, $AE_{L_2}$, $AE_{L_3}$. Through this process, $z_f$, $z_{l_2}$, $z_{l_3}$ are calculated, and the equation is as follows.

\begin{equation}
z_f = W_{e_f}x_f + b_{e_f}\label{eq13}
\end{equation}
\begin{equation}
z_{l_2} = W_{e_{l_2}}x_{l_2} + b_{e_{l_2}}\label{eq14}
\end{equation}
\begin{equation}
z_{l_3} = W_{e_{l_3}}x_{l_3} + b_{e_{l_3}}\label{eq15}
\end{equation}

The derived $z_f$, $z_{l_2}$, $z_{l_3}$ are inputted into the initial features $f_f$, $f_{l_2}$, $f_{l_3}$ and provided to the LSTM model.

\subsection{LSTM-based Network State Passing Process}
An MPNN updates the next hidden state of a node by receiving information from neighboring nodes. In the process of deriving the next state through information reception, an RNN-based model that processes input and output in sequential units can be used. Considering the gradual expansion of physical networks, LSTM is used in the proposed model to solve long-term dependency problems. First, to obtain the flow state of the reference node, $m_f$, the LSTM-based message function is inputted with $h_{l_2}$ and $h_{l_3}$, which contain connection information and information of neighboring nodes. At this point, the flow data is the data collected based on layer 3 devices. The derived $m_f$ data is stored in $m_{f,l_2}$ considering the L2 topology data. After completing the tasks related to all $l_2$ and $l_3$ relevant to the flow based on L2 and L3 topology data, the last $m_f$ is stored as the next $h_f$.

The $m_{f,l_2}$ obtained by the above procedure can be fed into LSTM along with $h_{l_2}$ to update the next hidden state of $l_2$. The derived $m_{l_2}^{t+1}$ is used to update the hidden state $h_{l_3}$ of all $l_3$ related to each $l_2$, based on the L2 and L3 topology data. The above three procedures are repeated $K$ times, and the resulting $h_f^T$, $h_{l_2}^T$, and $h_{l_3}^T$ are fed into the readout function, which is the final procedure for evaluating network performance.

\subsection{Network State Readout with Skip Connection}
With deep neural networks, there is a tendency for performance to improve as the network gets deeper. However, if it becomes too deep, the vanishing or exploding gradient problem can occur. To address this, short skip connections have been used to supplement the fully connected layer of the readout function that derives the predictions in the proposed method. If $x_p$ is the performance metric of the network to be evaluated, which is composed of $h_f^T$, $h_{l_2}^T$, and $h_{l_3}^T$, the equation is as follows:

\begin{equation}
y_p = R_p(x_p) = R'_p(x_p) + x_p\label{eq16}
\end{equation}

\RestyleAlgo{ruled}
\SetKwComment{Comment}{/* }{ */}
\SetKwInput{KwData}{Input}
\SetKwInput{KwResult}{Output}

\begin{figure}[htbp]
\centerline{\includegraphics[width=8.5cm]{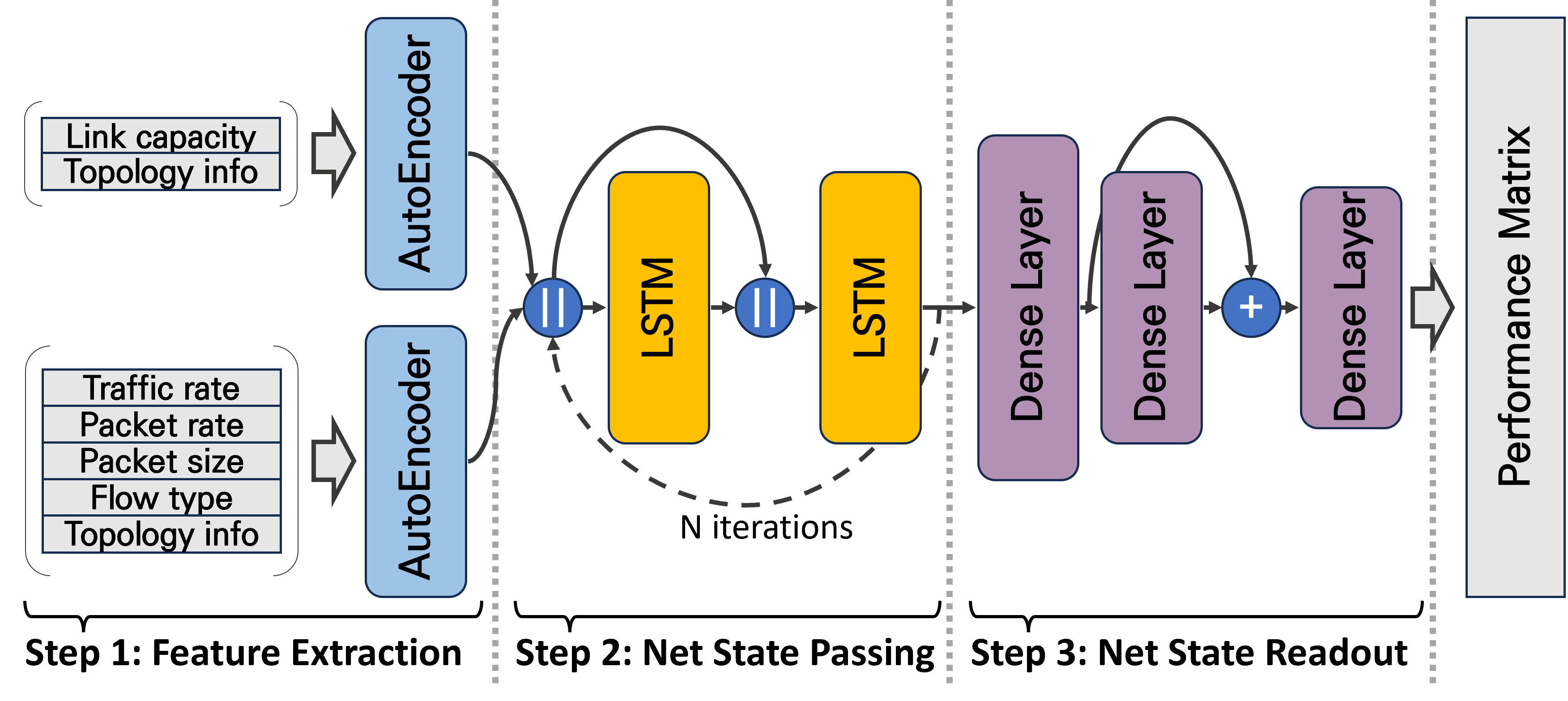}}
\caption{The architecture of purposed model.}
\label{fig4}
\end{figure}

\begin{algorithm}[hbt!]
\caption{Purposed Model Algorithm}\label{alg:two}
\KwData{$x_{f},x_{l_2},x_{l_3},F,L_2,L_3$}
\KwResult{$\hat{y}_{p}$}

\

\textbf{Step 1 :} Feature extraction

\lWhile{$f \in F$}{$z_f \gets AE_{f,l_3}(x_f,x_{t_3})$}
\lWhile{$l_2 \in L_2$}{$z_{l_2} \gets AE_{l_2}(x_{l_2})$}
\lWhile{$l_3 \in L_3$}{$z_{l_3} \gets AE_{f,l_3}(x_f,x_{t_3})$}

\

$(f_f^0, f_{l_2}^0, f_{l_3}^0) \gets (z_f, z_{l_2}, z_{l_3})$\;

\

\textbf{Step 2 :} Network state passing

\While{$k < K$}{

    \textbf{Step 2-1 :} aggregate all flow-specific $(l_2,l_3)$ link information.
    
    \While{$f \in F$}{
        \While{$(l_2,l_3) \in f$}{
            $m_f^k \gets M_f([h_f^k,h_{l_2}^k,h_{l_3}^k])$\;
            $m_{f,l_2}^{k+1} \gets m_f^k$\;
        }
        $h_f^{k+1} \gets m_f^k$\;
    }

    \textbf{Step 2-2 :} update network status based on flow-specific $l_2$ link.
    
    \While{$l_2 \in L_2$}{
        $h_{l_2}^{k+1} \gets U_{l_2}(h_{l_2}^k,m_{f,l_2}^{k+1})$\;
        \While{$l_3 \in l_2$}{
        $m_{f,l_2,l_3}^{k+1} \gets M_{l_2,l_3}([m_{f,l_2}^{k+1},h_{l_2}^{k+1}])$\;
        }
    }
    
    \textbf{Step 2-3 :} update network status based on flow-specific $l_2$-related $l_3$ link.
    
    \lWhile{${l_3} \in L_2$}{
        $h_{l_3}^{k+1} \gets U_{l_3}(h_{l_3}^k,m_{f,l_2,l_3}^{k+1})$
    }
    $k = k+1$\;
}

\

\textbf{Step 3 : } Network state readout

$\hat{y}_{p}=R_{p}(h_{f}^K,h_{l_2}^K,h_{l_3}^K)$\;
\end{algorithm}

\section{Performance Evaluation}
In this section, we compared the baseline model and its performance as provided in the ITU Challenge "ITU-ML5G-PS-007: Creating a network digital twin using real network data." 

\subsection{Experimental Setup}
The dataset used in the experiment was the DTN real datasets provided by UPC-BNN. The dataset is generated for 11 different topology scenarios in the DTN testbed configured as shown in Fig. 5, where each topology consists of randomly selected 5-8 nodes\cite{data}. Fig. 6 shows an example of a topology where the left topology (a) consists of 5 nodes and 5 bidirectional edges, and the right topology (b) consists of 8 nodes and 13 bidirectional edges.

\begin{figure}[htbp]
\centerline{\includegraphics[width=8.5cm]{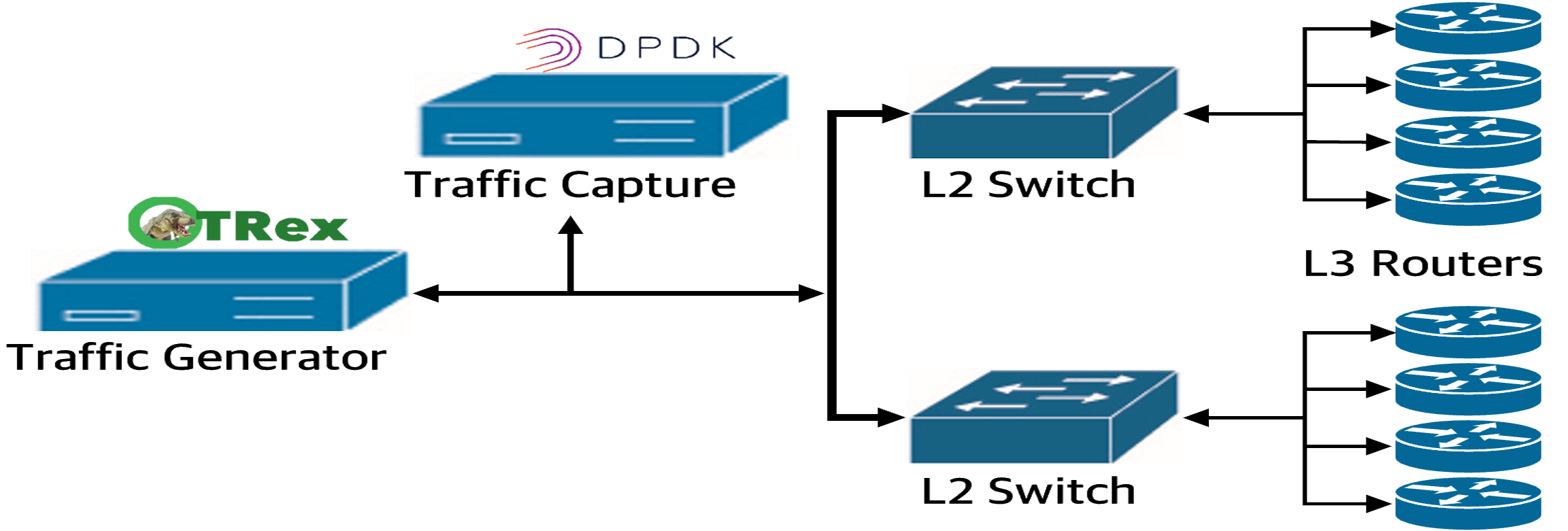}}
\caption{DTN testbed structure.}
\label{fign-1}
\end{figure}

\begin{figure}[htbp]
\centerline{\includegraphics[width=8.5cm]{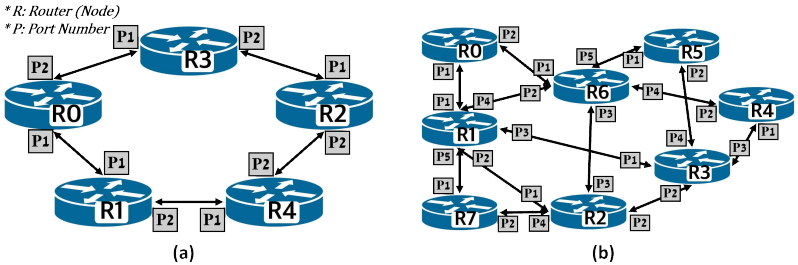}}
\caption{Network topology scenario example.}
\label{fign-2}
\end{figure}

The dataset was provided as part of the ITU AI Challenge\cite{itu challenge}. In this paper, the MPNN model used connection information between nodes contained in L2 and L3 topology data, as well as link capacity, traffic rate, packet rate, packet size, and flow type for learning. This model used network performance data (e.g., average delay, jitter, packet loss) in the training data because it performs the task of predicting network performance data. Specifically, in this experiment, the target was set as the average delay per src-dst. The ranges and units of the data used in the experiment are shown in Table 1.

\begin{table}[htbp]
\caption{The training datasets.}
\begin{center}
\begin{tabular}{c|cc}
\hline
\textbf{Data}   & \multicolumn{2}{c}{\textbf{Information}}  \\ \hline \hline
Link Capacity   & \multicolumn{2}{c}{1.0 / 10.0 / 80.0 (Gbps)}    \\ \hline
Topology   & \multicolumn{1}{c|}{Number of Nodes}  & 5$\sim$8  \\ \cline{2-3}
Component   & \multicolumn{1}{c|}{Number of Edges}  & 10$\sim$26   \\ \hline
Traffic Rate   & \multicolumn{2}{c}{8.58447072e+05 $\sim$3.23802572e+08 (bits/sec)}   \\ \hline
Packet Rate   & \multicolumn{2}{c}{221.7 $\sim$43856.35 (packets/sec)}    \\ \hline
Packet Size   & \multicolumn{2}{c}{824.0 $\sim$11552.0 (bits)}    \\ \hline
Flow Type   & \multicolumn{2}{c}{CBR(Constant-Bit Rate) / MB(Multi-burst)}  \\ \hline
\end{tabular}
\end{center}
\end{table}

For performance metrics in evaluating the prediction results, MAPE(Mean Absolute Percentage Error) was used for the training set, and MAPE, MAE(Mean Absolute Error), MSE(Mean Squared Error), and MSLE(Mean Squared Log Error)  were mainly used to evaluate the time series prediction model for the test set.

\subsection{Results and Discussion}
The proposed model was divided into four main models for experimentation. The first model is the AE-MPNN model, which combines AE and MPNN. The second model is the AE-SMPN model, which adds skip-connections to the readout function of MPNN. The AE-SMPN model was further divided into AE-SMPN2, AE-SMPN3, AE-SMPN4 models based on the number of layers in the readout function. Table 2 displays the hyperparameters of each model.

\begin{table}[htbp]
\caption{Hyperparameters for the proposed model.}
\begin{center}
\begin{tabular}{c|c}
\hline
\textbf{Hyperparameters} & \textbf{Value}   \\ \hline
Epoch                    & 50   \\
AE Embedding Size        & 64   \\
MPNN Iterations          & 8    \\
Learning Rate            & 0.001    \\
Activation Function      & SELU \\
Optimizer                & Adam \\
Loss Fuction             & MAPE \\
Evaluation Metrics      & MAPE / MAE / MSE / MSLE \\ \hline
\end{tabular}
\end{center}
\end{table}

The loss graph of the proposed model is shown in Fig. 7 and Fig. 8. In this case, Adam was used as the optimization function, and MAPE was used as the loss function and the results are shown in Table 3.

\begin{figure}[htbp]
\centerline{\includegraphics[width=8.5cm]{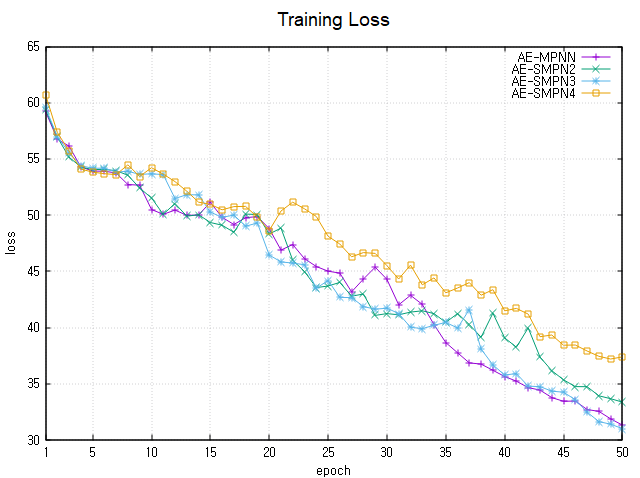}}
\caption{Training loss of proposed models.}
\label{fig5}
\end{figure}

\begin{figure}[htbp]
\centerline{\includegraphics[width=8.5cm]{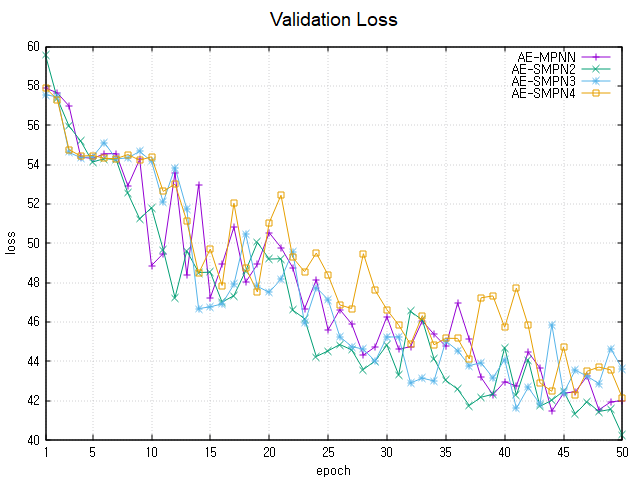}}
\caption{Validation loss of proposed models.}
\label{fig6}
\end{figure}

\begin{table}[htbp]
\caption{Performance of each model}
\begin{center}
\begin{tabular}{c|c|c|c|c|c|c}
\hline
\textbf{Models}& \multicolumn{3}{c|}{\textbf{MAPE(\%)}}& \textbf{MAE}& \textbf{MSE}& \textbf{MSLE} \\
\cline{2-7}
                        & \textbf{\textit{train}}& \textbf{\textit{val}}& \textbf{\textit{test}}& \textbf{\textit{test}}& \textbf{\textit{test}}& \textbf{\textit{test}} \\
\hline
Baseline & - & 44.29 & 100.52 & - & - & - \\
AE-MPNN & \textbf{31.32} & 42.01 & \textbf{42.42} & \textbf{4.05} & \textbf{92.11} & \textbf{0.75} \\
AE-SMPN2 & 33.37 & \textbf{40.23} & 47.17 & 4.84 & 130.64 & 1.27 \\
AE-SMPN3 & \textbf{31.01} & 43.60 & \textbf{42.82} & \textbf{4.20} & \textbf{109.92} & \textbf{0.77} \\
AE-SMPN4 & 35.08 & \textbf{41.99} & 46.35 & 5.26 & 197.30 & 1.03 \\
\hline
\end{tabular}
\label{tab1}
\end{center}
\end{table}

\section{Conclusion}
With the emergence and proliferation of new forms of large-scale services such as smart homes, virtual reality/augmented reality, the increasingly complex networks are raising concerns about significant operational costs. As a result, the need for network management automation is emphasized, and DTN technology is expected to become the foundation technology for autonomous networks.
In this paper, we proposed a network evaluation models for DTN, called AE-MPNN and AE-SMPN, which can help to achieve automated networks based on user intent. The model has been trained and validated using real network data, focusing on addressing two issues in the existing SOTA model, RouteNet. As part of the ITU Challenge "ITU-ML5G-PS-007: Creating a Network Digital Twin with Real Network Data," we compared the performance of our proposed model with the baseline model provided in the challenge. Through experimentation and validation, we confirmed that the proposed model achieves high accuracy in network evaluation tasks. However, it has the limitation that it may add potential overhead to the embedding process based on the autoencoder, so we want to do further research on model lightweighting and also study model modularization for integration into the pipeline. The proposed mechanism is expected to enable the automation of network management to achieve user intent and help reduce average latency, thereby providing improved services in large-scale network environments.

\section*{Acknowledgment}
This work was partly supported by Innovative Human Resource Development for Local Intellectualization program through the Institute of Information \& Communications Technology Planning \& Evaluation(IITP) grant funded by the Korea government(MSIT) (IITP-2024-RS-2022-00156287, 50\%) and This work was supported by Institute of Information \& communications Technology Planning \& Evaluation (IITP) grant funded by the Korea government(MSIT) (No.2021-0-02068, Artificial Intelligence Innovation Hub, 50\%)

\vspace{12pt}

\end{document}